\documentclass{article}
\pdfoutput=1

\PassOptionsToPackage{numbers}{natbib}
\usepackage[final]{neurips_mic_2021}

\usepackage[utf8]{inputenc} %
\usepackage[T1]{fontenc}    %
\usepackage{hyperref}       %
\usepackage{url}            %
\usepackage{booktabs}       %
\usepackage{inconsolata}
\usepackage{amsmath}
\usepackage{amsfonts}       %
\usepackage{nicefrac}       %
\usepackage{microtype}      %
\usepackage{xcolor}         %
\usepackage{cleveref}
\usepackage{subcaption}
\usepackage{wrapfig}
\usepackage{graphicx}
\usepackage{float}

\usepackage{algorithm}
\usepackage{algpseudocode}

\title{Efficient Pragmatic Program Synthesis with Informative Specifications}

\author{%
  Saujas Vaduguru \\
  IIIT Hyderabad \\
  \small{\texttt{saujas.vaduguru@research.iiit.ac.in}} \\
   \And
   Kevin Ellis \\
   Cornell University \\
   \small{\texttt{kellis@cornell.edu}} \\
   \And
   Yewen Pu \\
   Autodesk Research \\
   \small{\texttt{yewen.pu@autodesk.com}} \\
}

\begin{document}

\maketitle

\begin{abstract}
Providing examples is one of the most common way for end-users to interact with program synthesizers.
However, program synthesis systems assume that examples consistent with the program are chosen at random, and do not exploit the fact that users choose examples pragmatically. Prior work \cite{pu2020pragmatic} modeled program synthesis as pragmatic communication, but required an inefficient enumeration of the entire program space. In this paper, we show that it is possible to build a program synthesizer that is both pragmatic \emph{and} efficient by approximating the joint distribution of programs with a product of independent factors, and performing pragmatic inference on each factor separately. 
This factored distribution approximates the exact joint distribution well when the examples are given pragmatically, and is compatible with a basic neuro-symbolic program synthesis algorithm. Surprisingly, we find that the synthesizer assuming a factored approximation performs \emph{better} than a synthesizer assuming an exact joint distribution when evaluated on natural human inputs. This suggests that humans may be assuming a factored distribution while communicating programs.
\end{abstract}

\section{Introduction}

Program synthesizers are systems that take a specification of user intent as input, and synthesize a program in a \emph{domain-specific language} (DSL) that satisfies the specification. Providing examples is the standard form of specification for end users of state-of-the-art program synthesizers, as examples are both intuitive for humans to provide and explicitly checkable by machines. However, program synthesizers typically assume the examples are chosen randomly \cite{REPL,RobustFill,FlashFill,Sketch}, and don't leverage the fact that humans choose examples pragmatically to convey their intent \cite{RationalPedagogical}. As a consequence, existing synthesizers are both \emph{inefficient} -- having to reason over a complex space of programs conditioned on \emph{random} specifications, and \emph{unintuitive} -- not treating the given specification pragmatically.

Prior work \citep{pu2020pragmatic} posed program synthesis as a reference game, allowing for the application of the Rational Speech Acts (RSA) model \citep{frank2012predicting} towards pragmatic program synthesis. They have shown that by treating user given examples pragmatically, one can build synthesizers that are more intuitive to use and require fewer examples when compared to the non-pragmatic ones. However, the formulation in \cite{pu2020pragmatic} considers each program as an atom, and requires multiple enumerations over the entire program space. This does not scale to combinatorially complex program spaces.
Thus, one is left with a dilemma: One can either make a synthesizer pragmatic at the cost of extreme inefficiency, or make it efficient and scalable to a large space of programs at the cost of it being non-pragmatic.

We take a step towards making a program synthesizer that is both pragmatic and efficient by factorizing the distribution of programs as a set of independent production rules as prescribed by the DSL's grammar. This factorization completely disregards the complicated correlative structures between these production rules neccesary to faithfully model the space of programs given the specification. 

Indeed, under a \emph{literal} specification that doesn't actively constrain the number of consistent programs, this naive factorization will generate programs that fail to satisfy the specification nearly every time -- it is a bad approximation. However, under a \emph{pragmatic} specification that selects a smaller number of consistent programs and is thus more \emph{informative}, the factorization preserves the sparsity of the joint distribution, and performs on par with a literal listener that explicitly models the exact joint distribution at a fraction of the inference cost. On the human speaker data of \cite{pu2020pragmatic}, the factored literal listener performs even \emph{better} than its exact counterpart, suggesting that humans may intuitively be assuming a factored distribution while communicating to the synthesizers. 

The factored representation also makes it efficient to perform the recursive reasoning required for RSA algorithm, as each factor is a distribution over a small number of concepts. We find that the factored pragmatic synthesizer achieves reasonably good performance when compared to its exact counterpart. More importantly, our factorized approach allows for the introduction of learned (i.e. neural symbolic) program synthesizers, which makes it applicable to large program spaces.

\section{Efficient Pragmatic Synthesis}
As in \citep{pu2020pragmatic}, we model program synthesis as a reference game between two agents -- a speaker $S$ and a listener $L$. The speaker chooses a specification -- a set of examples $D = u_1 u_2 \ldots u_n$ -- to communicate a program $h$ to the listener. In keeping with the literature on the Rational Speech Acts model, we also refer to an example as an \emph{utterance}. The communication is successful if the listener is able to infer, or \emph{synthesize}, the correct program given the speaker's utterances.

\subsection{Exact Pragmatic Program Synthesis}

In this communication game setting, the task of synthesis is to model the listener distribution $P_L(h|D)$ of programs given utterances. To model a \emph{pragmatic} listener, \citep{pu2020pragmatic} propose using the Rational Speech Acts (RSA) model \citep{frank2012predicting} to recursively reason about a speaker generating utterances according to a speaker distribution $P_S(D|h)$ of utterances given a program.

\begin{wrapfigure}{r}{0.45\textwidth}
\vspace{-5mm}
\begin{small}
\begin{align}
    & P_{L_0}(h|D) = \frac{l(h, D)}{\sum_{h'} l(h', D)} \label{global_l0} \\
    & P^{spec}_{S_1}(D|h) = \prod_{i = 1}^{n} P^{utt}_{S_1}(u_i|h, u_{1}^{i - 1}) \label{global_s1_spec} \\
    & P^{utt}_{S_1}(u_i|h, u_{1}^{i - 1}) = \frac{P_{L_0}(h|u_{1}^{i - 1}, u_i)}{\sum_{u'_i} P_{L_0}(h|u_{1}^{i - 1}, u'_i)} \label{global_s1_utt} \\
    & P_{L_1}(h|D) = \frac{P^{spec}_{S_1}(D|h)}{\sum_{h'} P^{spec}_{S_1}(D|h')} \label{global_l1}
\end{align}
\end{small}
\vspace{-5mm}
\end{wrapfigure}

The \emph{literal listener} $L_0$ reasons about the \emph{lexicon} $l(h, D)$. 
The lexicon function takes the value $1$ if the program $h$ is consistent with a specification $D$, and $0$ otherwise. The literal listener is defined in \Cref{global_l0}.
The \emph{pragmatic speaker} $S_1$ reasons about a literal listener. Since modelling a distribution over all specifications is intractable, \citep{pu2020pragmatic} factorize the speaker distribution over specifications autoregressively into a product of distributions over utterances (\cref{global_s1_spec,global_s1_utt}).
The \emph{pragmatic listener} $L_1$ reasons about $S_1$ to pragmatically synthesize programs (\cref{global_l1}).

Under this formulation, the distribution $P(h|D)$ is over the space of all programs, 
making pragmatically considering all possible alternatives intractable outside of simple domains.

\subsection{Efficient Synthesis with a Mean-field Approximation}

\begin{figure}
    \centering
    \begin{subfigure}{0.495\textwidth}
        \centering
        \includegraphics[width=0.7\textwidth]{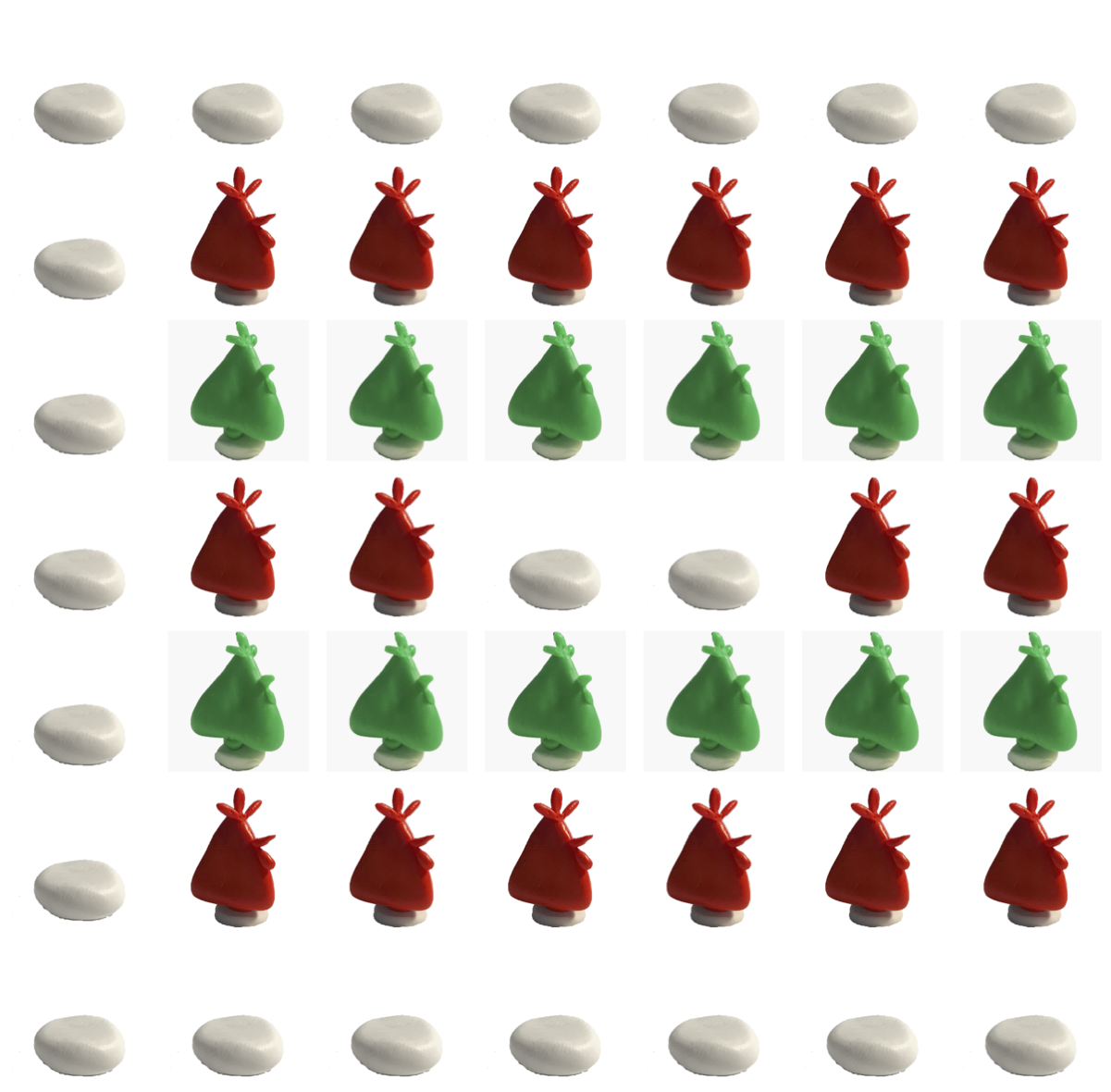}
        \caption{{\small $[\cdot, \cdot, \framebox{\texttt{1}}, \texttt{5}, \texttt{1}, \texttt{6}, \texttt{2}, \framebox{\texttt{chicken}}, \texttt{pebble}, \cdot, \framebox{\texttt{x}}, \lambda\texttt{z:z\%2}]$}}
        \label{fig:ex_prog1}
    \end{subfigure}
    \begin{subfigure}{0.495\textwidth}
        \centering
        \includegraphics[width=0.7\textwidth]{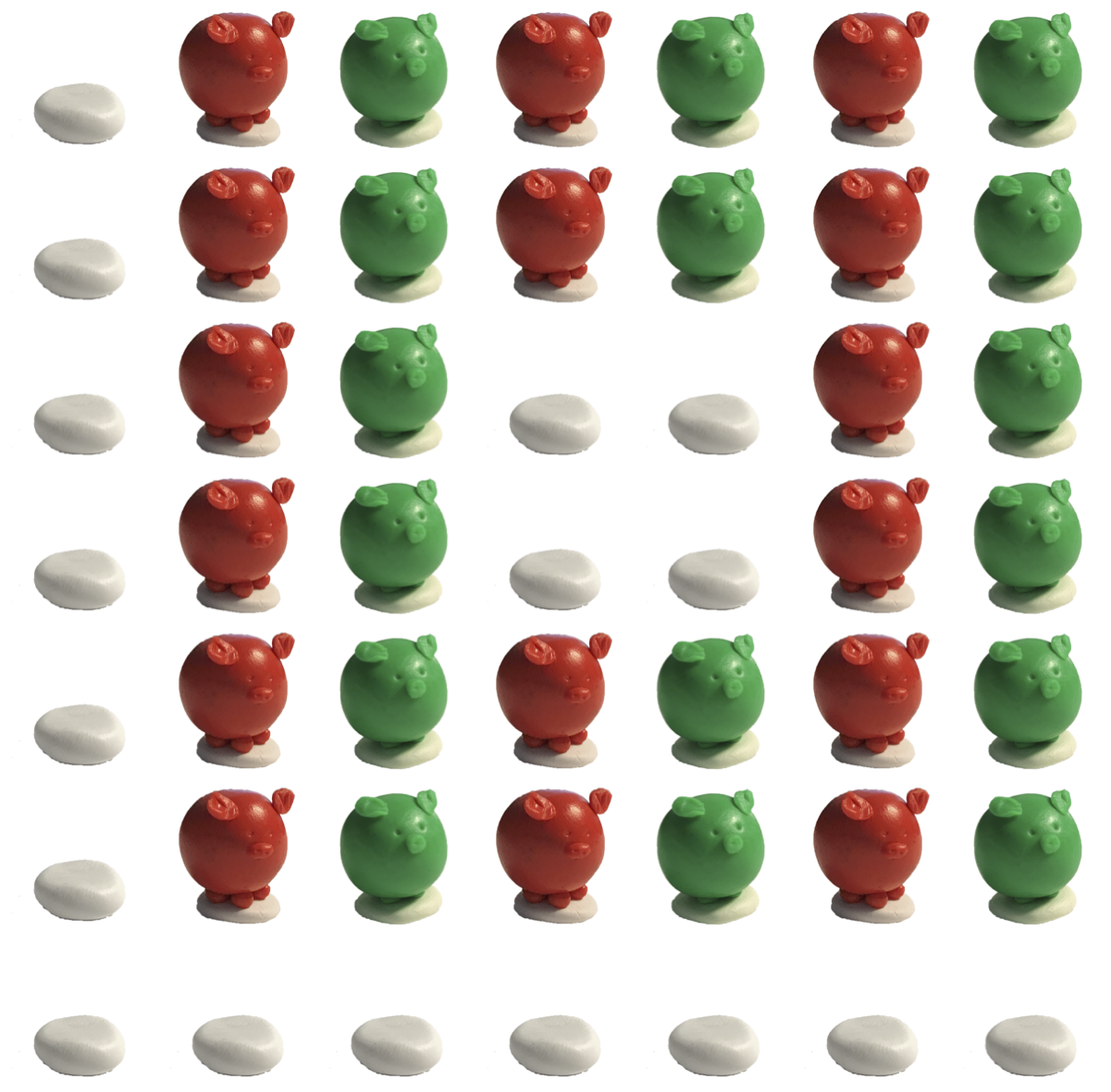}
        \caption{{\small $[\cdot, \cdot, \framebox{\texttt{0}}, \texttt{5}, \texttt{1}, \texttt{6}, \texttt{2}, \framebox{\texttt{pig}}, \texttt{pebble}, \cdot, \framebox{\texttt{y}}, \lambda\texttt{z:z\%2}]$}}
        \label{fig:ex_prog2}
    \end{subfigure}
    \caption{Two patterns in our layout domain and their corresponding programs, represented as a sequence of production rules: [\texttt{Program}, \texttt{Shape}, \texttt{Left}, \texttt{Right}, \texttt{Top}, \texttt{Bottom}, \texttt{Thickness}, \texttt{O}, \texttt{I}, \texttt{Colour}, \texttt{A1}, \texttt{A2}]. The symbol $\cdot$ indicates rules which only have 1 choice of expansion (\texttt{Program}, \texttt{Shape}, and \texttt{Colour}). The rules where these two programs differ are marked with a \framebox{box}.}
    \label{fig:ex_progs}
\end{figure}

Instead of viewing the program as an atomic referent, we view a program as a finite sequence of derivations in the grammar of the DSL. The grammar defines a set of $K$ non-terminals, and a number of rules $N_i \to \alpha_i^1, \ldots, N_i \to \alpha_i^n$ that expand $N_i$. Each step of the derivation selects a rule $N_i \to \alpha_i^j$, which expands a non-terminal $N_i$. We let $R_i$ denote the rule $N_i \to \alpha_i^j$ that expands the non-terminal $N_i$ in a program. When the production rules are not mutually 
recursive, the program can be simply represented as \emph{a finite sequence of production values}: $[R_1, R_2, \ldots, R_K]$. \Cref{fig:grammar} shows the specific DSL we use in our experiments and \Cref{fig:ex_progs} shows two sample programs from the DSL, represented as a finite sequence of productions.

Under this representation of the program, we can use a \emph{mean-field approximation} \citep[Ch.~10]{bishop} to naively factorize the distribution of programs $P(h|D)$ as
\begin{align*}
    Q(h|D) = Q^1(R_1|D)Q^2(R_2|D)\cdots Q^K(R_K|D)
\end{align*}
where $Q^i(N_i \to \alpha_i|D)$ is a distribution over rules that expand the non-terminal $N_i$. This approximation allows us to reason over each non-terminal independently.  \Cref{sec:marginal_proof} shows that minimizing the forward KL distance between $P(h|D)$ and $Q(h|D)$ amounts to performing supervised learning on the marginal distribution of $P$ on each factors $Q^i$ separately.

Pragmatic synthesis under the mean-field approximation amounts to carrying out the recursive computation for each productions independently (tractable as each $Q^i$ operates over a small number of productions), similar to \Cref{global_l0,global_s1_utt,global_s1_spec,global_l1}. This is shown \Cref{fact_l0,fact_s1,fact_l1}.
$Q^{i, spec}_{S_1}$ is defined autoregressively using $Q^{i, utt}_{S_1}$ as in $P^{spec}_{S_1}$. 

\begin{wrapfigure}{r}{0.5\textwidth}
\vspace{-5mm}
\begin{small}
\begin{align}
    & l^i(R_i, D) = \frac{\sum_{h: R_i \in h} l(h, D)}{\sum_{h} l(h, D)} \label{fact_lex} \\
    & Q^i_{L_0}(R_i|D) = \frac{l^i(R_i, D)}{\sum_{R'_i} l^i(R'_i, D)} \label{fact_l0} \\
    & Q^{i, utt}_{S_1}(u_j |R_i, u_{1}^{j - 1}) = \frac{Q^i_{L_0}(R_i|u_{1}^{j - 1}, u_j)}{\sum_{u'_j} Q^i_{L_0}(R_i|u_{1}^{j - 1}, u'_j)} \label{fact_s1} \\
    & Q^i_{L_1}(R_i|D) = \frac{Q^{i, spec}_{S_1}(D|R_i)}{\sum_{R'_i} Q^{i, spec}_{S_1}(D|R'_i)} \label{fact_l1}
\end{align}
\end{small}
\vspace{-5mm}
\end{wrapfigure}

The lexicon $l^i$ for the $i^{th}$ rule (\cref{fact_lex}) 
represents the fraction of all programs that satisfy the specification where the $i^{th}$ non-terminal is expanded using the rule $R_i$. We can use this notion of the lexicon to compute the factorized literal listener as a marginal over all other factors, which is the best approximation under the factorization.
Alternatively, the distribution $Q^i_{L_0}$ can be parametrized using a neural network trained to approximate the literal listener distribution. Given $Q$, we enumerate programs in decreasing order of probability until we find a consistent program, or run out of the search budget (\Cref{sec:appendix_search}).

\section{Experiments}

As an exploratory study, we evaluate our approach on a simple layout domain \citep{pu2020pragmatic}, where exact inference over the space of programs is possible. In this domain, a program is a pattern on a grid formed from a set of objects. These objects may be a colourless pebble, or a chicken or pig that may be red, green or blue. 
Each utterance reveals the object at one square on the grid, and the speaker has to communicate the pattern by revealing squares on the grid. The pattern is formed according to rules specified in the domain-specific language in \Cref{fig:grammar}. The specification is a sequence of objects, specified by the coordinates of the object on the grid, and its shape and colour.

\begin{figure}[h]
\centering
\begin{small}
\begin{align*}
    \texttt{Program} \to &\  \langle \texttt{Shape, Colour}\rangle \\
    \texttt{Shape} \to &\  \texttt{Box(Left, Right, Top, Bottom, Thickness, Outside, Inside)} \\
    \texttt{Left} \to &\  \texttt{0 | 1 | 2 | 3 | ... | 6} \\
    \texttt{Right} \to &\  \texttt{0 | 1 | 2 | 3 | ... | 6} \\
    \texttt{Top} \to &\  \texttt{0 | 1 | 2 | 3 | ... | 6} \\
    \texttt{Bottom} \to &\  \texttt{0 | 1 | 2 | 3 | ... | 6} \\
    \texttt{Thickness} \to &\  \texttt{1 | 2 | 3} \\
    \texttt{O}   \to &\  \texttt{chicken | pig} \\
    \texttt{I}   \to &\  \texttt{chicken | pig | pebble} \\
    \texttt{Colour}   \to &\  \texttt{[red , green , blue][A}_2\texttt{(A}_1\texttt{)]} \\
    \texttt{A}_1 \to &\  \texttt{x | y | x + y} \\
    \texttt{A}_2 \to &\ \lambda \texttt{z:0} \texttt{|} \lambda \texttt{z:1} \texttt{|} \lambda \texttt{z:2} \texttt{|} \lambda \texttt{z:z\%2} \texttt{|} \lambda \texttt{z:z\%2+1} \texttt{|} \lambda \texttt{z:2*(z\%2)} \\
\end{align*}
\end{small}
\vspace{-8mm}
\caption{Grammar of the DSL}
\label{fig:grammar}
\end{figure}

\subsection{The Models}
We consider 6 different listener models -- literal and pragmatic variants of 3 different types of models. The first type is the models from \citep{pu2020pragmatic}, which enumerate the entire program space during inference. These are the \emph{joint literal listener} $L_{J_0}$ and the \emph{joint pragmatic listener} $L_{J_1}$, since they model the full joint distribution from \cref{global_l0,global_l1} respectively. 
The second type is based on the factorized listener distributions from \cref{fact_l0,fact_l1}. In this type, the literal listener distributions are obtained by enumerating over the space of programs.
These are termed the \emph{factorized literal listener} $L_{F_0}$ and the \emph{factorized pragmatic listener} $L_{F_1}$. 
The third type of model is also based on the factorized listener distributions, but the literal listener distribution
is approximated using a neural network (\Cref{sec:neural}). These are termed the \emph{neural literal listener} $L_{N_0}$ and the \emph{neural pragmatic listener} $L_{N_1}$.

\begin{figure}[b]
    \centering
    \begin{subfigure}{0.49\textwidth}
        \centering
        \includegraphics[width=\textwidth]{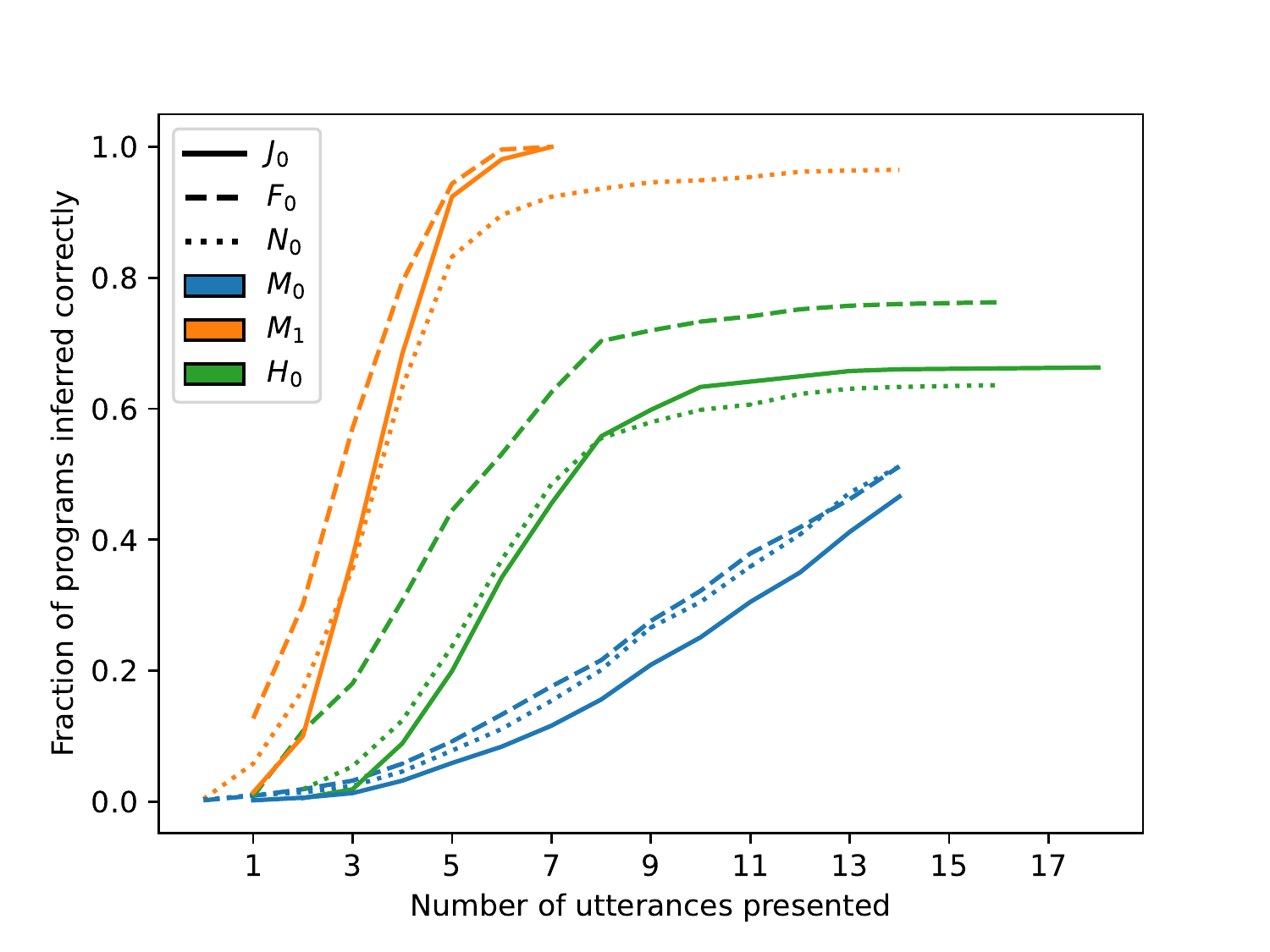}
        \caption{Literal listeners $L_{J_0}$, $L_{F_0}$, and $L_{N_0}$}
        \label{fig:l0_results}
    \end{subfigure}
    \begin{subfigure}{0.49\textwidth}
        \centering
        \includegraphics[width=\textwidth]{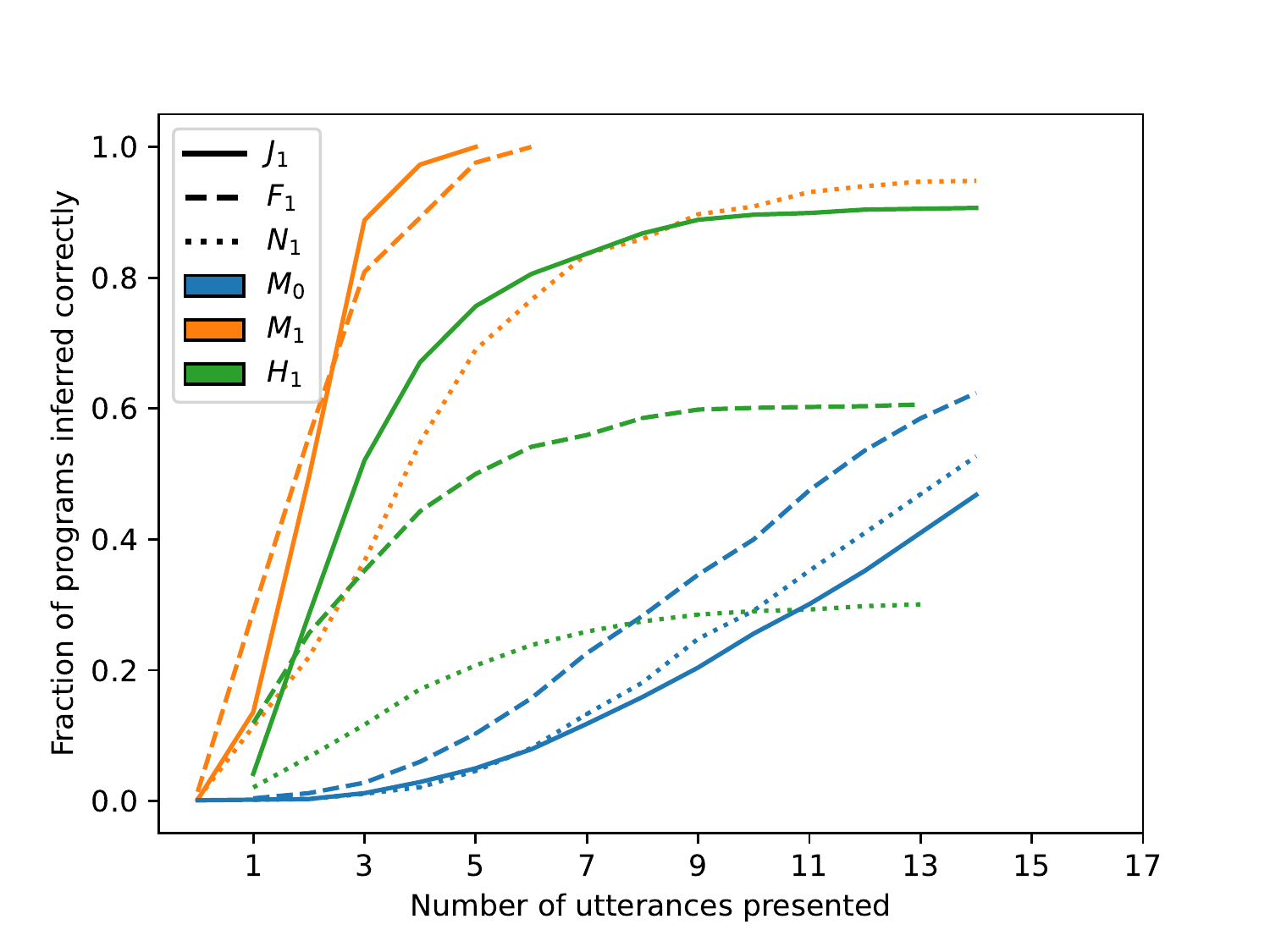}
        \caption{Pragmatic listeners $L_{J_1}$, $L_{F_1}$, and $L_{N_1}$}
        \label{fig:l1_results}
    \end{subfigure}
    \caption{Comparison of different listeners (represented using different line types), literal: $L_{J_0}$, $L_{F_0}$, $L_{N_0}$ and pragmatic $L_{J_1}$, $L_{F_1}$, $L_{N_1}$, interacting with different speakers (represented using different line colors) $S_{M_0}$, $S_{H_0}$, $S_{H_1}$ and $S_{M_1}$. X-axis shows number of utterances presented (i.e. user efforts) and Y-axis shows fraction of programs inferred correctly (i.e. communication accuracy)}
    \label{fig:results}
\end{figure}

\subsection{The Data}
We use speaker models to generate specifications from 1000 randomly chosen programs from the DSL. We use a \emph{literal speaker} $S_{M_0}$ to randomly generate utterances that are true of the program, and a \emph{pragmatic speaker} $S_{M_1}$ to generate utterances under the joint speaker distribution in \cref{global_s1_spec}. These are idealized cases of a perfectly literal and pragmatic (according to RSA) listeners respectively. 

We leverage human speaker data from \cite{pu2020pragmatic}, collected by interactions between $\texttt{Human} \rightarrow L_{J_0}$ and $\texttt{Human} \rightarrow L_{J_1}$. We denote a human speaker interacting with $L_{J_0}$ as $S_{H_0}$, and a human speaker interacting with $L_{J_1}$ as $S_{H_1}$. Note that even when interacting with $L_{J_0}$, a human speaker still behaves pragmatically (i.e. $S_{H_0} \neq S_{M_0}$). Details of how the data are processed are presented in \Cref{sec:data_appendix}.

\begin{figure}
    \centering
    \begin{subfigure}{0.2\textwidth}
    \begin{subfigure}{0.98\textwidth}
        \centering
        \includegraphics[width=\textwidth]{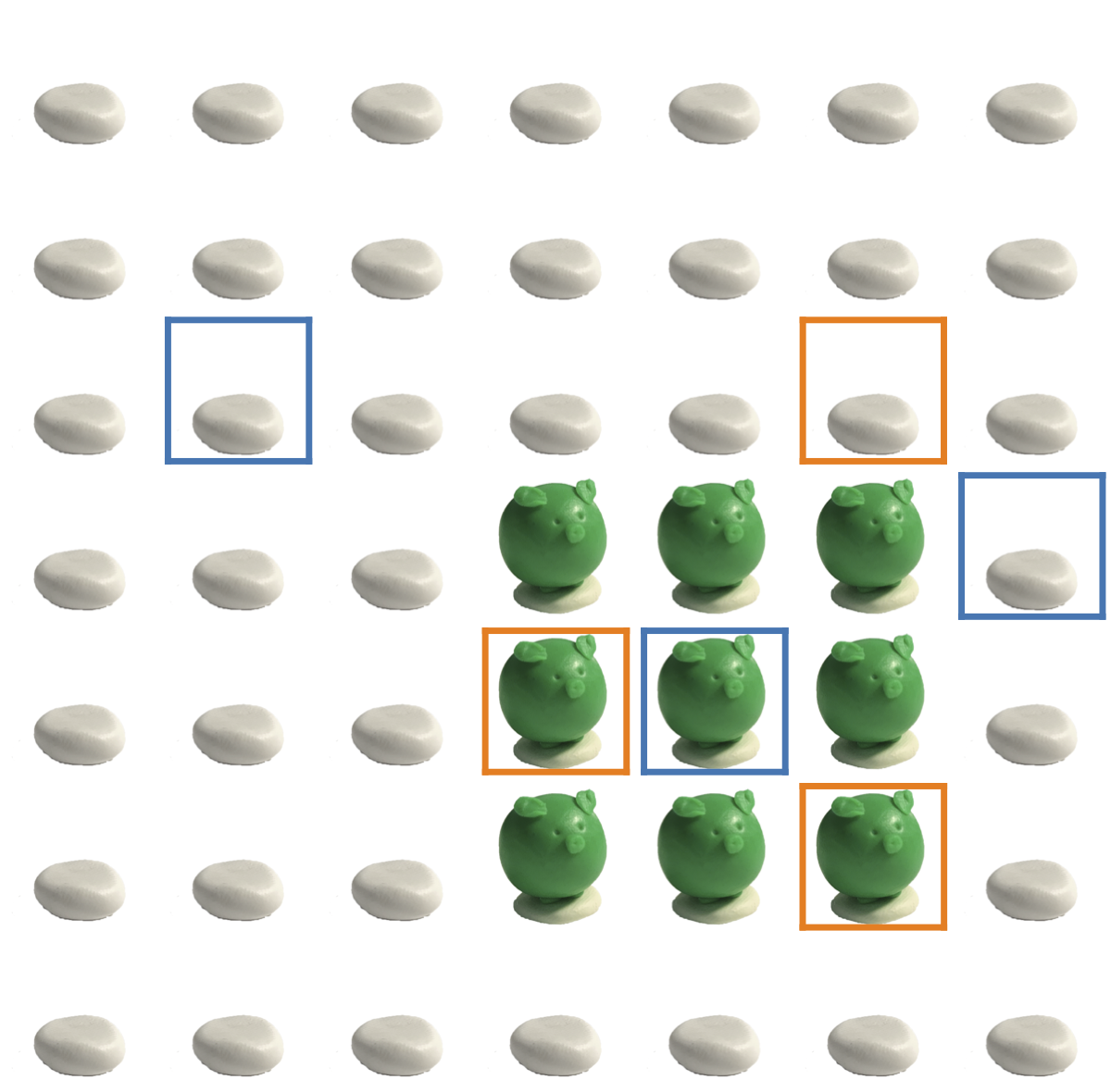}
        \label{fig:marginal_spec}
    \end{subfigure}
    
    \begin{subfigure}{0.98\textwidth}
        \centering
        \includegraphics[width=\textwidth]{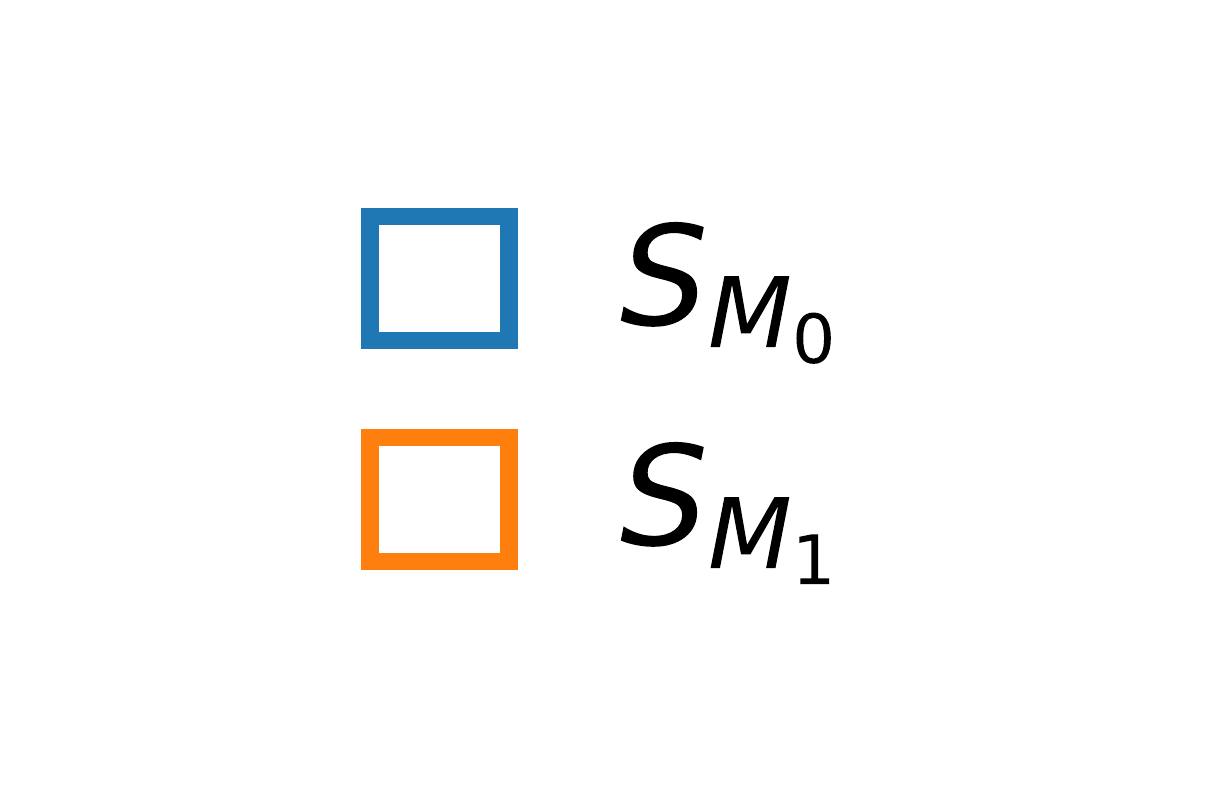}
    \end{subfigure}
    \end{subfigure}
    \begin{subfigure}{0.79\textwidth}
    \begin{subfigure}{0.49\textwidth}
        \centering
        \includegraphics[width=\textwidth]{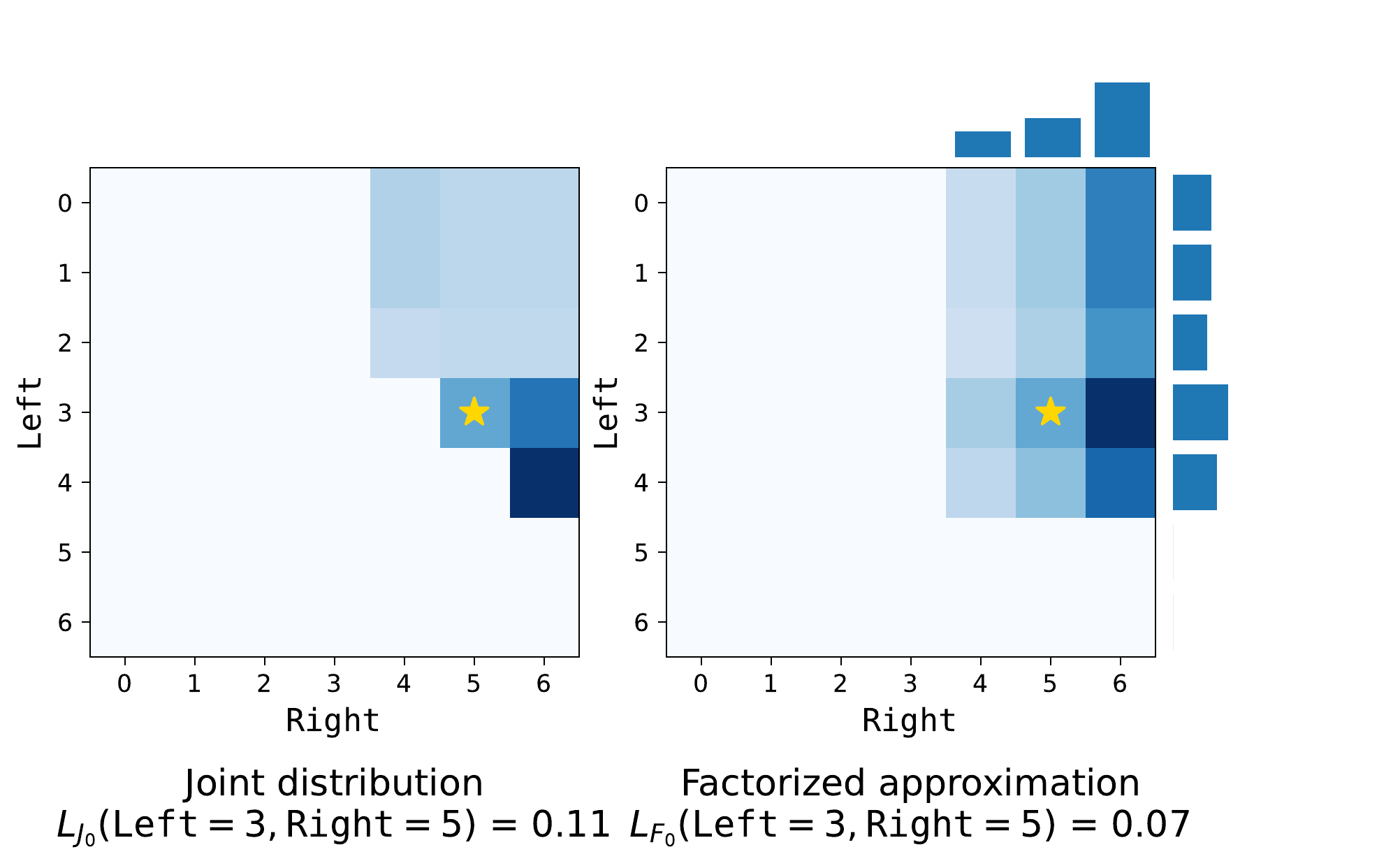}
        \caption{$S_{M_0} \rightarrow L_{J_0}, S_{M_0} \rightarrow L_{F_0}$}
        \label{fig:marginal_s0_l0}
    \end{subfigure}
    \begin{subfigure}{0.49\textwidth}
        \centering
        \includegraphics[width=\textwidth]{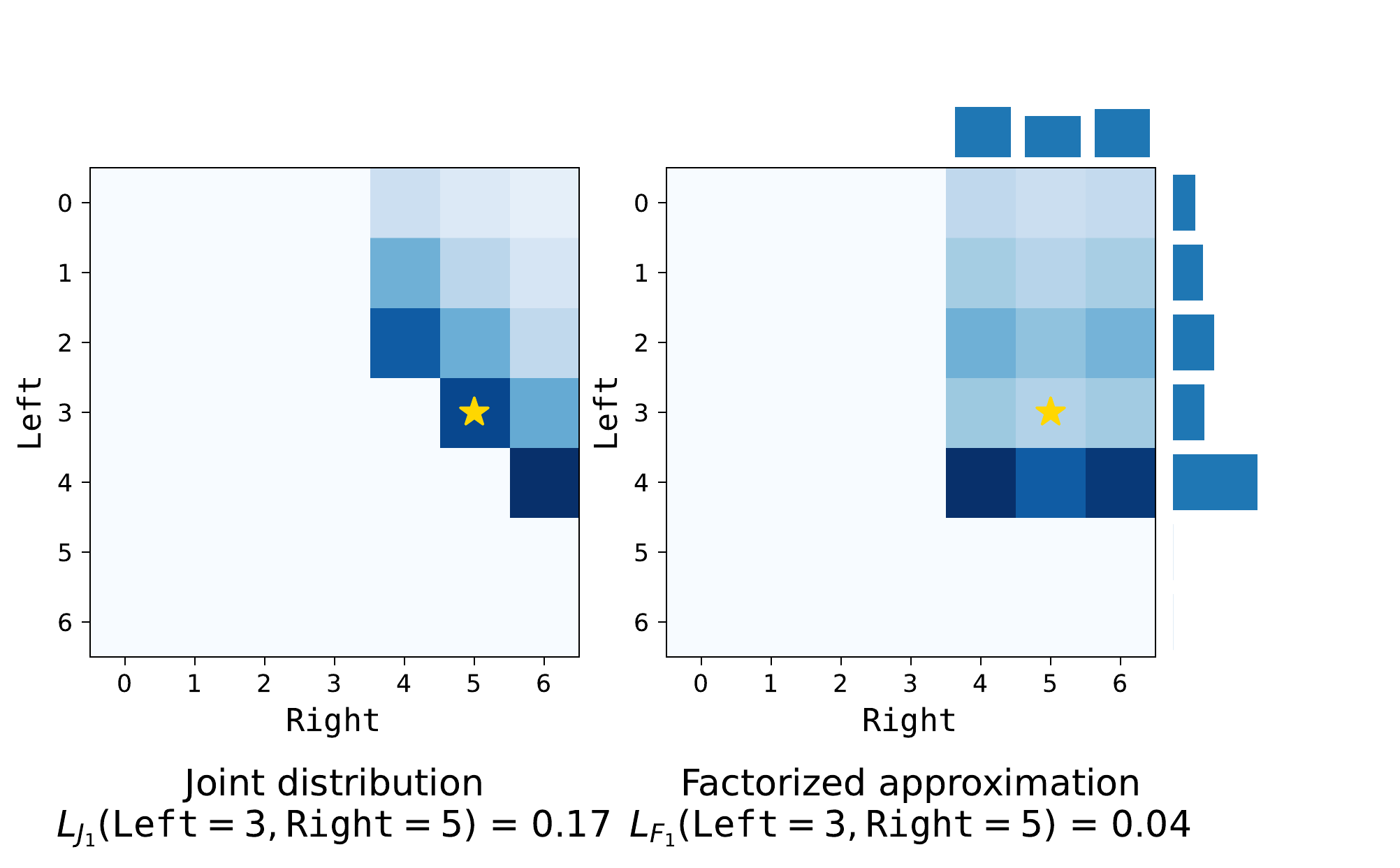}
        \caption{$S_{M_0} \rightarrow L_{J_1}, S_{M_0} \rightarrow L_{F_1}$}
        \label{fig:marginal_s0_l1}
    \end{subfigure}

    \begin{subfigure}{0.49\textwidth}
        \centering
        \includegraphics[width=\textwidth]{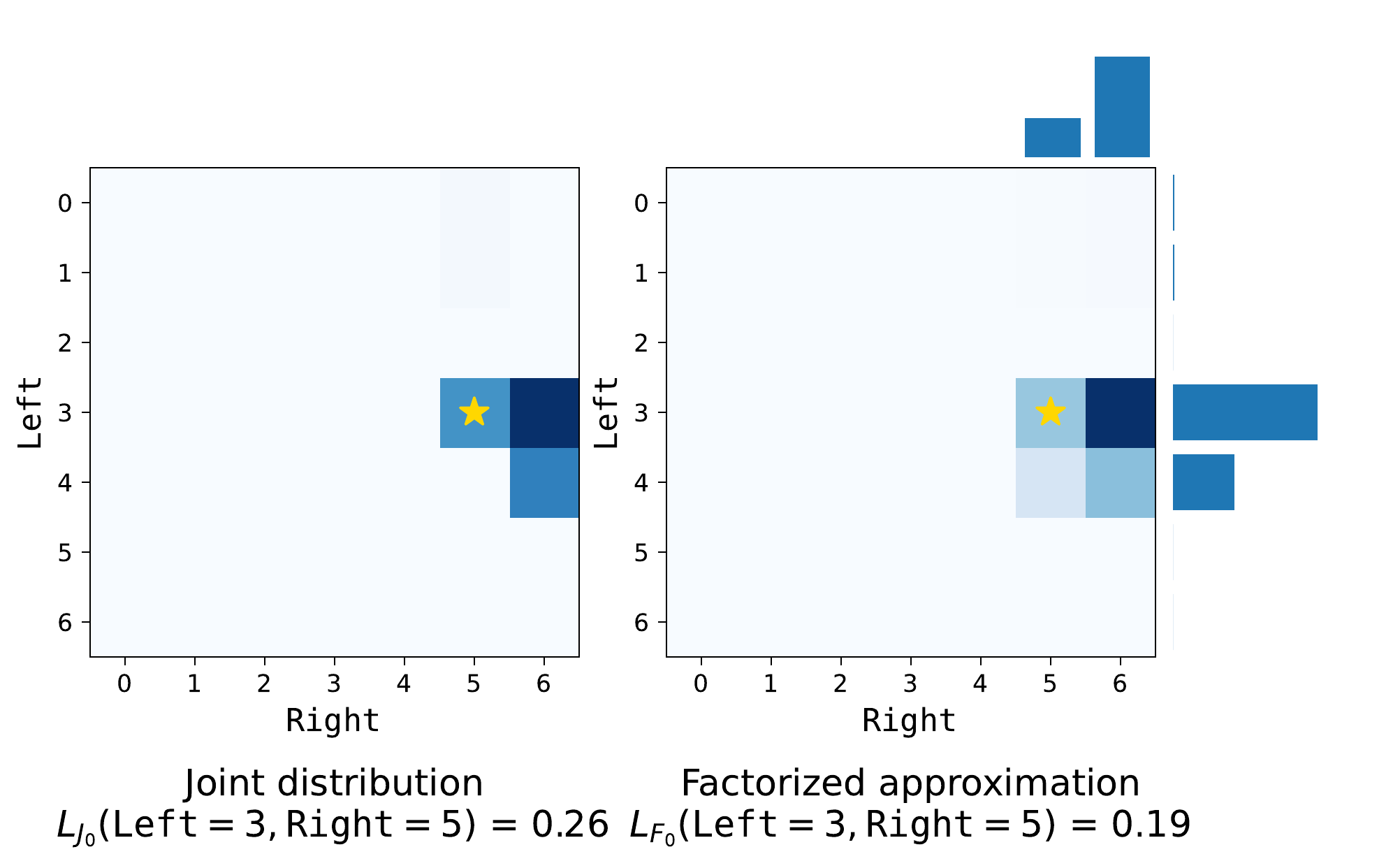}
        \caption{$S_{M_1} \rightarrow L_{J_0}, S_{M_1} \rightarrow L_{F_0}$}
        \label{fig:marginal_s1_l0}
    \end{subfigure}
    \begin{subfigure}{0.49\textwidth}
        \centering
        \includegraphics[width=\textwidth]{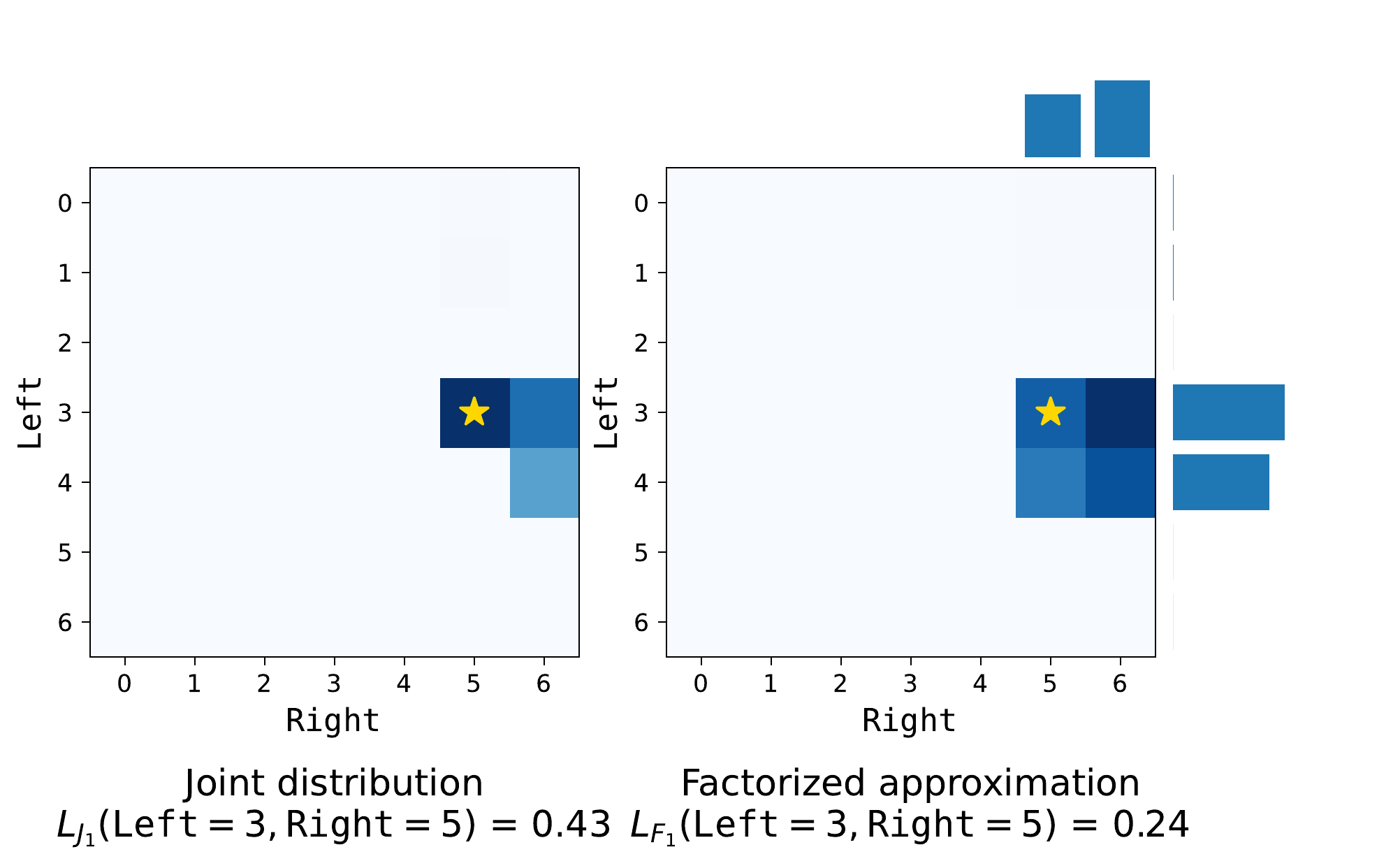}
        \caption{$S_{M_1} \rightarrow L_{J_1}, S_{M_1} \rightarrow L_{F_1}$}
        \label{fig:marginal_s1_l1}
    \end{subfigure}
    \end{subfigure}

    \caption{Heatmaps showing the distribution over the rules for two different non-terminals in the DSL -- \texttt{Left} and \texttt{Right} -- that specify the left and right edges of the pattern respectively, under the specification shown on the left. Each case shows the distribution produced by one type of listener (literal $L_{J_0}/L_{F_0}$ or pragmatic $L_{J_1}/L_{F_1}$) under a specification produced by a speaker model. In each case, the heatmap on the left shows the joint distribution over these two productions under the a joint listener model. The heatmap on the right shows the joint distribution over these two productions under a factorized model, with the distribution of each factor shown as a histogram. The rules that satisfy the program are indicated with a star.}
    \label{fig:marginals}
\end{figure}

\subsection{Results}

In \Cref{fig:l0_results}, we note that an idealized pragmatic speaker $S_{M_1}$ allows all the literal listener models -- $L_{J_0}$, $L_{F_0}$, and $L_{N_0}$ -- to perform well. This suggests that under pragmatic specifications from $S_{M_1}$, the factorized distribution of $L_{F_0}$ approximates the exact, joint distribution $L_{J_0}$ well, due to informative specification concentrating the probability mass in a smaller region that's amendable to a factorized model. We illustrate this effect on a concrete instance in \Cref{fig:marginals}. This affirms our intuition that a pragmatic specification should make inference \emph{easier}, and allows for an extremely simple neuro-symbolic synthesizer. As expected, performance on uninformative specifications from $S_{M_0}$ is poor for both literal and pragmatic models.

Surprisingly, the factored literal listener $L_{F_0}$ performs \emph{better} than the exact model $L_{J_0}$ when given human generated specifications of $S_{H_0}$. This is suprising as $S_{H_0}$ is expressly collected in \cite{pu2020pragmatic} from the \emph{interactive} pair $S_{H_0} \rightarrow L_{J_0}$. This suggests that humans may intuitively be assuming a factored distribution while communicating with the synthesizer, for instance, choosing examples to communicate a particular cooridnates of one of the edges of the pattern.

For the pragmatic listeners in \Cref{fig:l1_results}, we note that the factored model $L_{F_1}$ approximates the exact model $L_{J_1}$ well when given idealized specifications from $S_{M_1}$, i.e. the communicative pair $S_{M_1} \rightarrow L_{F_1}$ performs as well as $S_{M_1} \rightarrow L_{J_1}$. However, using human-generated specifications $S_{H_1}$, we find that the pair $S_{H_1} \rightarrow L_{F_1}$ performs worse than $S_{H_1} \rightarrow L_{J_1}$ . We suspect this is due to an ``interaction gap'' where data from $S_{H_1}$ was collected by \cite{pu2020pragmatic} during interaction between the pair $S_{H_1} \rightarrow L_{J_1}$. The human speaker had the opportunity to adjust their strategy and adapt specifically to $L_{J_1}$, but could not dynamically adapt to $L_{F_1}$ in our experiments.

We also note that the neural model $L_{N_1}$ performs worse when compared to $L_{J_1}$ and $L_{F_1}$. We suspect this is due to approximation errors of the neural model accumulate when different factors are combined (via multiplication), and during RSA's recursive reasoning process. This results in larger gaps between $L_{N_1}$ and $L_{F_1}$ than between $L_{N_0}$ and $L_{F_0}$.

\section{Discussion and Future Directions}

We find that a synthesizer using a naive factorized distribution over programs performs well, but only when given informative, pragmatic specifications such as those a generated by a human. One explanation is that informative specifications concentrating the probability mass over a smaller number of programs, as illustrated in \Cref{fig:marginals}.  Another explanation is that our DSL factors the hypothesis space similarly to how a human would factor it. This results in a communication scheme that focuses on conveying each factor independently, without significant loss of information. Formally characterizing this notion of ``concentrating the probability mass'' (perhaps in an information theoretic language) and conducting user studies exploring how humans conceptualize and reason about large, compositional space of programs are exciting directions for future work.

We also observe that our current neural pragmatic model $L_{N_1}$ performs poorly. We postulate that one can obtain better results by (i) having a better model and training schemes for approximating \cref{fact_l0} and (ii) the opportunity to interact directly with users.

\bibliographystyle{unsrt}
\bibliography{neurips_2021.bib}

\begin{thebibliography}{1}

\bibitem{pu2020pragmatic}
Yewen Pu, Kevin Ellis, Marta Kryven, Josh Tenenbaum, and Armando Solar-Lezama.
\newblock Program synthesis with pragmatic communication.
\newblock In H.~Larochelle, M.~Ranzato, R.~Hadsell, M.~F. Balcan, and H.~Lin,
  editors, {\em Advances in Neural Information Processing Systems}, volume~33,
  pages 13249--13259. Curran Associates, Inc., 2020.

\bibitem{REPL}
Kevin Ellis, Maxwell Nye, Yewen Pu, Felix Sosa, Josh Tenenbaum, and Armando
  Solar-Lezama.
\newblock Write, execute, assess: Program synthesis with a repl.
\newblock In H.~Wallach, H.~Larochelle, A.~Beygelzimer, F.~d\textquotesingle
  Alch\'{e}-Buc, E.~Fox, and R.~Garnett, editors, {\em Advances in Neural
  Information Processing Systems}, volume~32. Curran Associates, Inc., 2019.

\bibitem{RobustFill}
Jacob Devlin, Jonathan Uesato, Surya Bhupatiraju, Rishabh Singh, Abdel-rahman
  Mohamed, and Pushmeet Kohli.
\newblock Robustfill: Neural program learning under noisy i/o.
\newblock In {\em Proceedings of the 34th International Conference on Machine
  Learning - Volume 70}, ICML'17, page 990–998. JMLR.org, 2017.

\bibitem{FlashFill}
Sumit Gulwani.
\newblock Automating string processing in spreadsheets using input-output
  examples.
\newblock {\em SIGPLAN Not.}, 46(1):317–330, January 2011.

\bibitem{Sketch}
Armando Solar{-}Lezama, Liviu Tancau, Rastislav Bod{\'{\i}}k, Sanjit~A. Seshia,
  and Vijay~A. Saraswat.
\newblock Combinatorial sketching for finite programs.
\newblock In John~Paul Shen and Margaret Martonosi, editors, {\em Proceedings
  of the 12th International Conference on Architectural Support for Programming
  Languages and Operating Systems, {ASPLOS} 2006, San Jose, CA, USA, October
  21-25, 2006}, pages 404--415. {ACM}, 2006.

\bibitem{RationalPedagogical}
Patrick Shafto, Noah~D. Goodman, and Thomas~L. Griffiths.
\newblock A rational account of pedagogical reasoning: Teaching by, and
  learning from, examples.
\newblock {\em Cognitive Psychology}, 71:55--89, 2014.

\bibitem{frank2012predicting}
Michael~C Frank and Noah~D Goodman.
\newblock Predicting pragmatic reasoning in language games.
\newblock {\em Science}, 336(6084):998--998, 2012.

\bibitem{bishop}
Christopher~M. Bishop.
\newblock {\em Pattern Recognition and Machine Learning (Information Science
  and Statistics)}.
\newblock Springer-Verlag, Berlin, Heidelberg, 2006.

\end{thebibliography}

\appendix

\section{Factorized Approximation of Joint Listener Distribution} \label{sec:marginal_proof}

For simplicity, let us consider a setting where $K = 2$. So, we have the joint distribution $P(h|D)$ approximated by $Q(h|D) = Q^1(R_1|D)Q^2(R_2|D)$.

We would like to minimize the quantity $\textit{KL}(P(R_1 R_2 | D) || Q(R_1 R_2 | D) )$.
\begin{align*}
    & \min_{Q} \textit{KL}(P(R_1 R_2|D)||Q(R_1 R_2|D)) \\
    & = \min_{Q^1, Q^2} \textit{KL}(P(R_1 R_2|D)||Q^1(R_1|D)Q^2(R_2|D)) \\
    & = \min_{Q^1, Q^2} \sum_{R_1, R_2} P(R_1 R_2|D) \log\left(\frac{P(R_1 R_2|D)}{Q^1(R_1|D)Q^2(R_2|D)}\right) \\ %
    & = \min_{Q^1, Q^2} \sum_{R_1, R_2} P(R_1 R_2|D) \log P(R_1 R_2|D) - \sum_{R_1, R_2} P(R_1 R_2|D) \log\left(Q^1(R_1|D)Q^2(R_2|D)\right) \\
    & = \min_{Q^1, Q^2} -\sum_{R_1, R_2} P(R_1 R_2|D) \log\left(Q^1(R_1|D)Q^2(R_2|D)\right) \\  %
    & = \max_{Q^1, Q^2} \sum_{R_1, R_2} P(R_1 R_2|D) \log\left(Q^1(R_1|D)Q^2(R_2|D)\right) \\
    & = \max_{Q^1} \sum_{R_1, R_2} P(R_1 R_2|D) \log Q^1(R_1|D) + \max_{Q^2} \sum_{R_1, R_2} P(R_1 R_2|D) \log Q^2(R_2|D)
\end{align*}

We can maximize each term of the sum separately. Without loss of generalization, let's focus on $Q^1$. 
\begin{align*}
    & \max_{Q^1} \sum_{R_1, R_2} P(R_1 R_2|D) \log Q^1(R_1|D) \\
    & = \max_{Q^1} \sum_{R_1} \sum_{R_2} P(R_1 R_2|D) \log Q^1(R_1|D) \\
    & = \max_{Q^1} \sum_{R_1} \log Q^1(R_1|D) \sum_{R_2} P(R_1 R_2|D) \\ 
    & = \max_{Q^1} \sum_{R_1} P(R_1|D) \log Q^1(R_1|D) \\
    & = \min_{Q^1} -\sum_{R_1} P(R_1|D) \log Q^1(R_1|D)
\end{align*}
We see that minimizing the KL divergence can be reduced to minimizing the cross-entropy loss between the factor of the mean-field approximation corresponding to a non-terminal symbol and the marginal of the joint distribution over all other non-terminals.

To train a learned model, we can sample programs and corresponding specfications and train with this objective,
\begin{align*}
    \min_{Q^1} \mathop{\mathbb{E}}_{r_1 \sim P(R_1|D)}[-\log Q^1(R_1|D)]
\end{align*}

\section{Search algorithm} \label{sec:appendix_search}

\begin{algorithm*}
\caption{Best-first search for programs} \label{alg:search}
\begin{algorithmic}
\Require Specification $D$, distributions over rules $Q^i(R_i|D)$, search budget $B$
\Ensure Program $\langle R_1, \ldots, R_K\rangle$ which satisfies $D$
\For{$i = 1$ to $K$}
    \State Sort rules $R_i^j$ in decreasing order of $Q^i(R_i^j|D)$
\EndFor
\State Searched set $S \gets \{\}$
\State Priority queue $P \gets []$ \Comment{Max priority queue with score $p$}
\State \texttt{enqueue}($P, \langle R_1^1, \ldots, R_K^1\rangle$) \Comment{$p = \sum_{i} \log Q^i(R_i^1|D)$}
\While{$|S| < B$}
\State $\langle R_1^{j_1}, \ldots, R_K^{j_K}\rangle \gets \texttt{dequeue}(L)$
\If{$\langle R_1^{j_1}, \ldots, R_K^{j_K}\rangle \in S$}
\State continue
\EndIf
\If{$\langle R_1^{j_1}, \ldots, R_K^{j_K}\rangle$ is consistent with $D$}
\State return $\langle R_1^{j_1}, \ldots, R_K^{j_K}\rangle$
\EndIf
\State $S \gets S \cup \{\langle R_1^{j_1}, \ldots, R_K^{j_K}\rangle\}$
\For{$i = 1$ to $K$}
    \If{replacing $R_i^{j_i}$ with $R_i^{j_i + 1}$ results in a valid program}
        \State \texttt{enqueue}($P, \langle R_1^{j_1}, \ldots, R_{i - 1}^{j_{i - 1}}, R_i^{j_i + 1}, R_{i + 1}^{j_{i + 1}}, \ldots R_K^{j_K}\rangle$) \Comment{$p = \sum_{i} \log Q^i(R_i^{j_i}|D)$}
    \EndIf
\EndFor
\EndWhile
\end{algorithmic}
\end{algorithm*}

\Cref{alg:search} describes the search algorithm. We use a best-first search algorithm, where we start with the highest probability program by considering the highest probability rule for each non-terminal. We then iterate over programs in decreasing order of probability. At each step, we replace one rule in the program with the next best alternative for that rule, and add it to a priority queue. We terminate the search when we find a consistent program, or exceed the search budget. For our experiments, we use a search budget of 50.

\section{Data processing} \label{sec:data_appendix}

\subsection{Speaker models}
We use the speaker models $S_{M_0}$ and $S_{M_1}$ to generate specifications according to an idealized literal and pragmatic speaker distribution respectively. The literal speaker model $S_{M_0}$ selects utterances that are true of the program with uniform probability. The pragmatic speaker model $M_1$ generates specifications autoregressively, at each step choosing the utterance with the highest probability according to \Cref{global_s1_utt}. For each program, we generate a specification of size 15 using each of the models $S_{M_0}$ and $S_{M_1}$ as a sequence of utterances $u_1, u_2, \ldots, u_n$. 

A listener model is presented with incrementally larger prefixes of this sequence of utterances, and the synthesis is terminated if the intended program is correctly identified by the listener model.

\subsection{User interaction data}
For each model $L_{J_0}$ and $L_{J_1}$, we have a set of trials collected by \cite{pu2020pragmatic}. We term a user interacting with $L_{J_0}$ as $S_{H_0}$ and a user interacting with $L_{J_1}$ as $L_{H_1}$. Each trial presents the user with a program in the DSL, and the user is taked with communicating the program to the model. The user reveals a sequence of utterances $u_1, u_2, \ldots, u_n$. 

We consider the utterances as a sequence, in the other they were provided by the user. We process the specifications to remove duplicate utterances within a trial, and retain only the first occurrence of the utterance. We have a total of 371 trials for $S_{H_0}$ and 386 trials for $S_{H_1}$.

The synthesis experiments are conducted in a similar manner to the experiments with data from speaker models.

\section{Neural models} \label{sec:neural}

We train a 2-hidden layer MLP, with each hidden layer having 256 units, as a neural literal listener $N_0$. The input to this model is a $7 \times 7 \times 6$ tensor $T$. For each element of the specification $(x, y, s, c)$, denoting the $x$-coordinate, $y$-coordinate, shape, and color (shape and color are mapped to an index between 0 and 2) respectively is encoded by setting $T[x, y, s] = 1$ and $T[x, y, 3 + c] = 1$. All other elements of the input are set to 0.

A common model for each production is trained for all productions. The model produces a $12 \times 7$ matrix $Q$, where $Q_{ij}$ represents the probability that a rule $R_j$ is chosen for the $i^{th}$ factor. The loss for each factor is computed as described in \Cref{sec:marginal_proof}, and summed to obtain the total loss for a sample.

The model is trained by sampling a batch of programs from a set of 10000 programs, sampling a specification using the literal speaker model $S_{M_0}$ for each program in the batch, and computing the loss between the predicted distribution for each factor and the ground truth value for factor. 

The model is trained with a batch size of 8 and specifications comprising of 2 to 25 elements for 150,000 steps.

\end{document}